\documentclass[11pt]{article}

\usepackage[final]{acl}

\usepackage{times}
\usepackage{latexsym}

\usepackage[T1]{fontenc}
\usepackage[utf8]{inputenc}

\usepackage{microtype}

\usepackage{inconsolata}

\usepackage{graphicx}
\usepackage{multirow}
\usepackage{amsmath}
\usepackage{amssymb}
\usepackage{bbm}
\usepackage{float}
\usepackage{booktabs}
\usepackage{enumitem}
\usepackage[para,online,flushleft]{threeparttable}
\usepackage{caption}
\usepackage{subcaption}
\usepackage{adjustbox}
\usepackage{colortbl}
\usepackage{tabularx}
\usepackage[font=footnotesize]{caption}

\usepackage{xcolor}
\newcommand{\ci}[2]{\textcolor{black!55}{(#1, #2)}}

\definecolor{ao(english)}{rgb}{0.0, 0.5, 0.0}
\definecolor{amaranth}{rgb}{0.9, 0.17, 0.31}
\newcolumntype{?}{!{\vrule width 1pt}}

\title{IUQ: Interrogative Uncertainty Quantification for Long-Form Large Language Model Generation}

\author{
  Haozhi Fan$^1$ ~~~~ Jinhao Duan$^{2*}$ ~~~~ Kaidi Xu$^3$\thanks{~~Corresponding to: Jinhao Duan <jinhao@cs.unc.edu> Kaidi Xu <kaidixu@cityu.edu.hk>.} \\
  $^1$University of Pennsylvania 
  $^2$UNC Chapel Hill \\
  $^3$City University of Hong Kong\\
}

\begin{document}
\maketitle
\begin{abstract}
Despite the rapid advancement of Large Language Models (LLMs), uncertainty quantification in LLM generation is a persistent challenge. Although recent approaches have achieved strong performance by restricting LLMs to produce short or constrained answer sets, many real-world applications require long-form and free-form text generation. A key difficulty in this setting is that LLMs often produce responses that are semantically coherent yet factually inaccurate, while the underlying semantics are multifaceted and the linguistic structure is complex. To tackle this challenge, this paper introduces \textbf{Interrogative Uncertainty Quantification (IUQ)}, a novel framework that leverages inter-sample consistency and intra-sample faithfulness to quantify the uncertainty in long-form LLM outputs. By utilizing an interrogate-then-respond paradigm, our method provides reliable measures of claim-level uncertainty and the model's faithfulness. Experimental results across diverse model families and model sizes demonstrate the superior performance of IUQ over two widely used long-form generation datasets. The code is available at \url{https://github.com/louisfanhz/IUQ}.
\end{abstract}

\section{Introduction}
\label{sec:introduction}

Large Language Models (LLMs) have shown remarkable improvement across a diverse range of Natural Language Processing tasks (\citealp{brown2020languagemodelsfewshotlearners}; \citealp{chowdhery2022palmscalinglanguagemodeling}; \citealp{kamalloo2023openDomainQA}). However, LLMs remain susceptible to hallucination, as they generate plausible answers that are factually incorrect (\citealp{zhang2023halluSurvey}; \citealp{huang2025halluSurvey}). 

\begin{figure}[t]
  \centering
  \includegraphics[width=0.95\columnwidth]{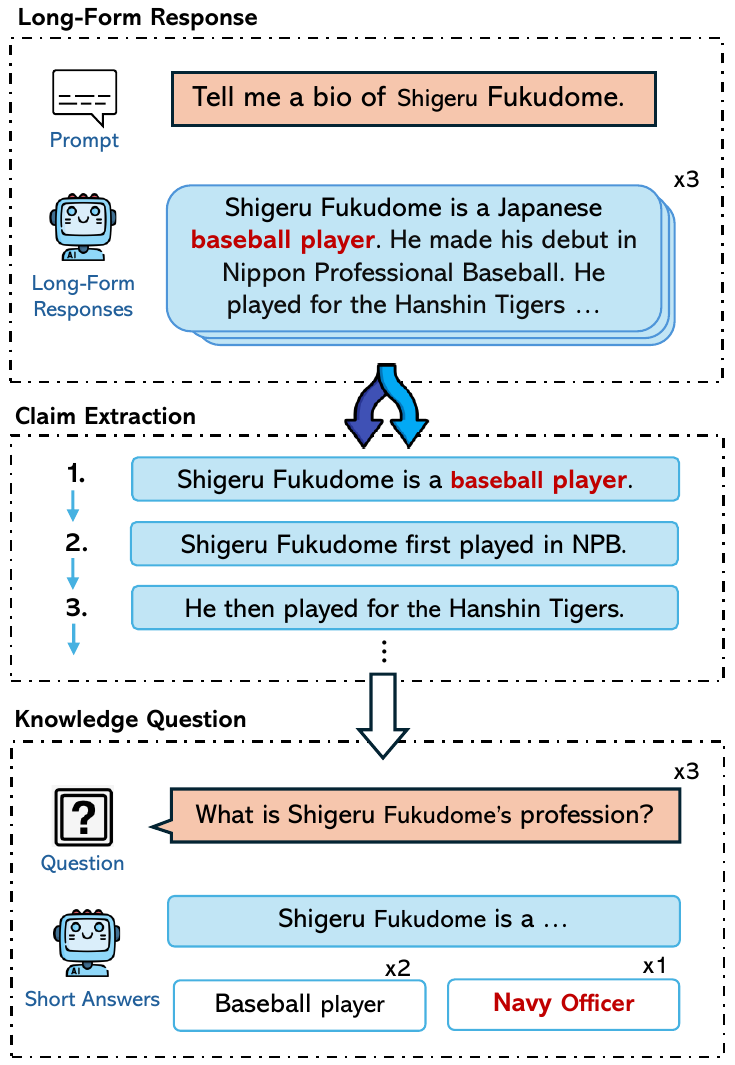}
  \caption{An example of LLM generation on biography. The model incorrectly identifies the individual's profession and fabricates a complete biography to maintain logical consistency. When multiple outputs are sampled, they exhibit a high level of similarity even though the biographies are based on false information. The model is then shown to be uncertain about the subject in a separate session asking specifically about the individual's profession.}
  \label{fig:intro_fig1}
  \vspace{-5mm}
\end{figure}

Recent Uncertainty Quantification (UQ) methods effectively measure hallucination within a confined answer space, where the models are prompted to generate short responses or answer multiple-choice questions (\citealp{kuhn2023semanticUncertainty}; \citealp{lin2024generatingConfidence}; \citealp{duan2024SAR}; \citealp{chen2024EigenScore}; \citealp{wang2024wordsequenceentropyuncertaintyestimation}). These approaches leverage token-probabilities or semantic entailment between responses to construct uncertainty estimates. However, in the scenario of free-form generation, where the response lengthens and exhibits structure and logic, it can be difficult to evaluate the entailment relationships between long answers, and aggregating token-probabilities becomes less indicative of uncertainty.

Current work on long-form UQ involves evaluating the semantic consistency between LLM responses. The long-form response is decomposed into sentences or claims, which are then compared against additional sampled responses to obtain uncertainty estimates (\citealp{manakul2023selfcheckgpt}; \citealp{zhang2024luq}; \citealp{jiang2024graphbased}; \citealp{wei2024longfact}). However, the following observation requires a closer look at the fine-grained contextual dependence in LLM generations: long-form output may differ across samples but is rarely self-contradictory due to next-token conditioning on the preceding context. Therefore, a critical challenge occurs when models fabricate information for the sake of logical consistency. As illustrated in \autoref{fig:intro_fig1}, when the LLM is prompted to provide information on a historical figure, it mistakenly identifies the individual's profession but continues to generate a plausible story to maintain logical consistency. On the other hand, when asked specifically about the individual's profession, the model returns inconsistent answers, showing uncertainty about the subject. The tendency for LLM to fabricate information is not captured by current UQ methods, which only evaluate consistency across sampled generations without testing the models' real knowledge on the subject.

Recent studies reveal LLMs hallucinate on facts that are present in the training data (\citealp{jiang2024onLLMHalluToKnownFacts}). When the topic is underrepresented, LLMs may even be overconfident about false knowledge (\citealp{kandpal2023longtailKnowledge}; \citealp{mallen2023whenNotToTrustLLM}; \citealp{ren2025investigateFactualKnowledgeBoundary}). These findings corroborates our observation that LLMs are not always faithful, and it has become increasingly difficult to identify incorrect information, as LLMs are more capable of formulating plausible responses (\citealp{hu2024towardFactualKnowledge}; \citealp{ji2024anah}).

To tackle this challenge, we first differentiate the claims in a generative context from the model's knowledge. Specifically, we incorporate an interrogator LLM to construct tailored short questions for each factual claim made in the long-form output. As a result, the original long-form response is decomposed into atomic knowledge decoupled from the generative context. To demonstrate correct knowledge, the model needs to answer consistently to each independent question and not provide answers that contradict the original claims. Naturally, the amount by which the model returns contradictory answers constitutes a measure of the tendency to fabricate information. 

We propose a novel UQ framework: Interrogative Uncertainty Quantification (IUQ) to facilitate fine-grained probing of long-form LLM responses. Different from other UQ methods that only implement inter-sample consistency, IUQ also enforces intra-sample claims consistency through independent question-answering. The strategy is analogous to an interrogate-respond scenario in which the responder is continuously questioned by the interrogator to screen untruthfulness.
Furthermore, since the short questions extracted from claims are independent of other sampled generations, IUQ presents a confidence landscape for each generation by viewing the quantified uncertainty of claims as data points in a time-series.

We evaluate IUQ on various model families of diverse sizes: GPT4o \citep{openai2024gpt4o}, Qwen2 \cite{yang2024qwen2}, Gemma-3 \citep{gemmateam2025gemma3}, Mistral \citep{jiang2023mistral7b}, LlaMA-3.1, and LlaMA-3.3~\cite{grattafiori2024llama3herdmodels}, with model sizes ranging from 24B up to 72B. We use two datasets tailored for long-form generation: FActScore \citep{min2023factscore}, which contains items of biography, and LongFact \citep{wei2024longfact}, which contains prompt sets on topics of art, science, and so on. Our contribution is the following:
\begin{itemize}
    \item We propose the Interrogative Uncertainty Quantification (IUQ) workflow that evaluates long-form responses through fine-grained probing of claim-level knowledge. Extensive experiments have demonstrated the effectiveness of IUQ over diverse model families.
    \item We highlight an under-explored phenomenon of long-form LLM generation where the models fabricate factual information to maintain logical consistency, and present a quantitative analysis of this tendency at the claim-level.
\end{itemize}

\begin{figure*}[t]
  \includegraphics[width=1.0\linewidth]{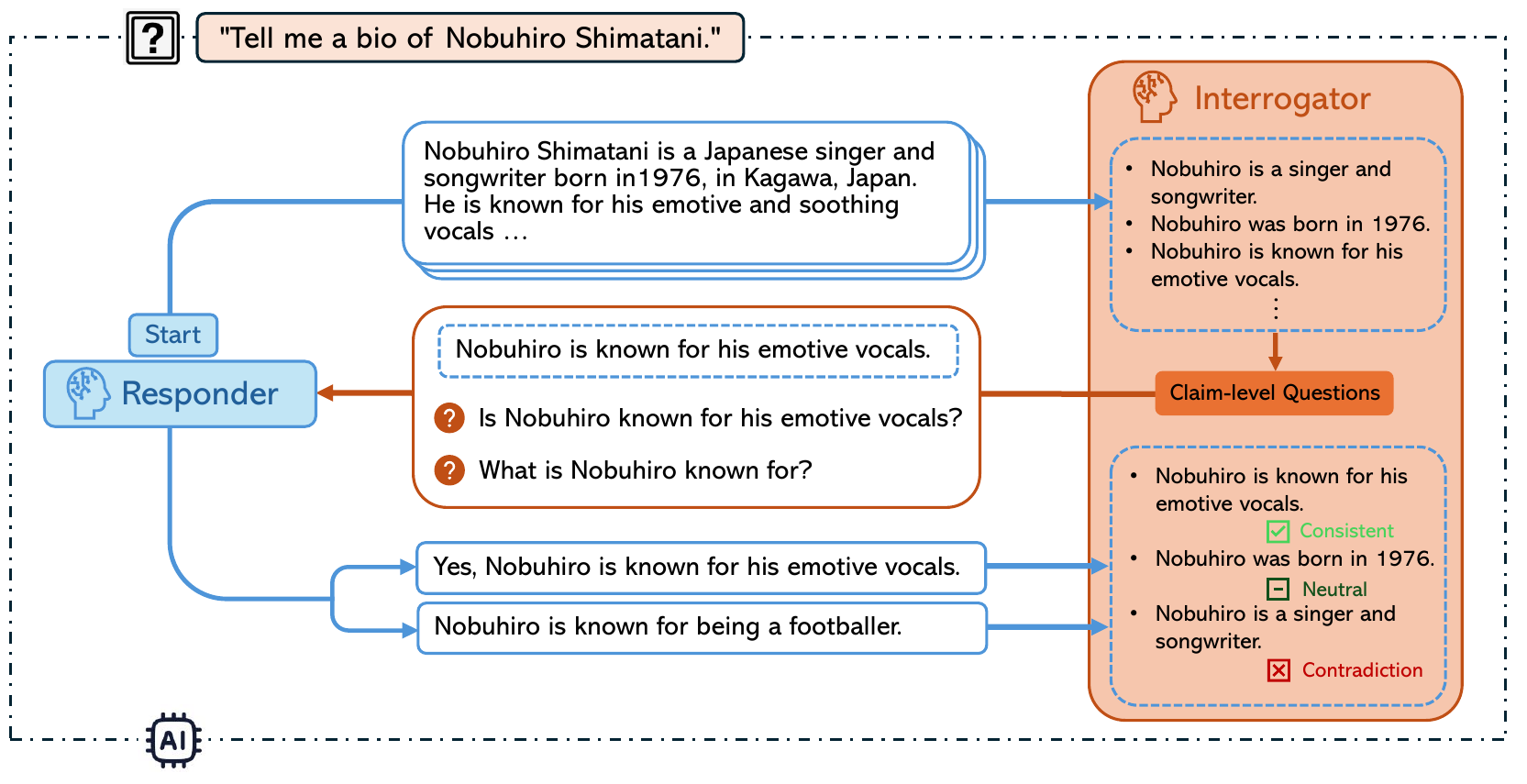} 
  \caption {The framework of Interrogative Uncertainty Quantification (IUQ): Responses are sampled from LLMs and decomposed into atomic claims. The model then answers questions that address the information contained in the claim in a separate session without context. The model is unfaithful about a claim if the claim contradicts the corresponding answers, which represent the model's knowledge on the subject.}
  \label{fig:pipeline}
\end{figure*}

\section{Related Work}

\noindent\textbf{Uncertainty Quantification} Existing UQ methods can be roughly categorized into white-box and black-box methods. White-box methods assume the model architecture is partially or completely visible (\citealp{kuhn2023semanticUncertainty}; \citealp{nikitin2024kernelEntropy}; \citealp{duan2024SAR}, \citealp{fadeeva2024factcheckingviaToken}; \citealp{duan2026truthprintmitigatinglargevisionlanguage}), whereas the black-box methods rely on the input prompts and LLM responses to measure uncertainties (\citealp{xiong2024canLLMexpressUncertainty}; \citealp{gao2024spuq}; \citealp{wang2025coinuncertaintyguardingselectivequestion}; \citealp{duan2025upropinvestigatinguncertaintypropagation}). Our work follows the line of black-box methods. Among them, \citet{tonolini2024bayesianPromptUQ} utilizes a weighted ensemble of semantically equivalent prompts to compute output uncertainty, where the weights are obtained through Bayesian variational inference. \citet{xiong2024canLLMexpressUncertainty} explores various strategies in prompting, sampling, and aggregating phases to acquire confidence scores from the model. \citet{gao2024spuq} perturbs the prompts and measures the semantic variation in responses. IUQ is distinct from these methods that it applies to long-form LLM outputs and decouples the contextual dependence to examine the models' real knowledge.

\noindent\textbf{Self-Consistency in LLMs} Self-consistency based approaches are proven to be effective in diverse domains associated with LLMs \citep{pan2024autoCorrectLLMSurvey}. \citet{wang2023selfconsistencyCoT} have shown significant improvement in Chain-of-thought prompting by sampling multiple paths and picking the most consistent answer. \citet{shinn2023reflexion} robustly induces better decision-making in various agentic tasks through linguistic feedback. For quantifying uncertainty, a general workflow for consistency estimation is to perform inter-sample consistency checks or let the models output verbal-confidence (\citealp{manakul2023selfcheckgpt}; \citealp{duan2023diffusionmodelsvulnerablemembership}; \citealp{chen2024universal}; \citealp{rivera2024combiningConfidenceElicitation}; \citealp{jiang2024graphbased}). \citet{kuhn2023semanticUncertainty} and \citet{lin2024generatingConfidence} utilize Natural Language Inference models and pairwise entailment to compute uncertainty estimates over a set of sampled responses. \citet{zhang2024luq} and \citet{jiang2024graphbased} utilize LLM to infer the supportiveness of responses to each claim. However, none of them address the intra-sample contextual dependence, which can lead to trusting fabricated yet consistent sampled outputs.

\section{IUQ: Interrogative Uncertainty Quantification}

IUQ focuses on fine-grained factuality and isolates atomic knowledge from the generative context. Structurally, IUQ is composed of a responder and an interrogator, with the interrogator continuously questioning the responder for the information it has generated, as shown in \autoref{fig:pipeline}. In practice, we use the same language model for both the responder and the interrogator to prevent systematic bias. Please refer to \autoref{appendix:prompts} for the prompts we used in IUQ.

\subsection{Response Generation}
Given a model $M$ and a prompt $x$, we sample N diverse responses from $M$ with temperature $T=t$. These responses comprise a set $\mathcal{R}$ such that $\mathcal{R}=\{R_1,\dots,R_N\}$, where $R_i=M_{T=t}(x)$ for $i\in\{1,\dots,N\}$. The generated responses are free-form texts that have variable lengths. The responses that refuse to answer are excluded (e.g. responses of "I don't know", "I cannot provide information").

\subsection{Claim-Level Question-Answering}
\label{sec:response_decomposition}
Unlike short-form outputs, the long-form generation of LLM is phrased in natural language consisting of syntax, factual information, and colloquial phrases. A common method to extract information from long text is to incorporate an LLM to decompose the generated text into the smallest possible semantic claims (\citealp{min2023factscore}; \citealp{song2024veriscore}; \citealp{jiang2024graphbased}). We follow the same practice and use the model $M$ to decompose the response $R$ to obtain a set of atomic claims $\mathcal{C}^{R}$, where
\begin{equation}
    \label{eqn:claims_set}
    \mathcal{C}^{R} = M_{T=0}(R, x) = \{c_1, c_2,\dots,c_k\},
\end{equation}

\noindent and $k$ is the number of claims returned by model $M$.

As discussed in \autoref{sec:introduction}, we aim to examine the model's knowledge by decoupling the claims from the generative context. This is achieved by utilizing LLM to extract the information contained in a claim and phrasing it as a question. We obtain the set of questions sampled for claim $c\in\mathcal{C}^R$ as
\begin{equation}
    \mathcal{Q}_{c} = M_{T=t}(c, x) = \{q_1, q_2,\dots,q_{n_q}\}, 
\end{equation}
where $n_q$ denotes the number of questions generated for claim $c$. We generate $\mathcal{Q}_c$ in a single inference request and require $n_q\leq3$ to prevent repetitive questions and limit computation costs.

Ideally, only one question should be derived for each claim due to its semantic atomicity; however, since the decomposition of the original response could be non-exhaustive (e.g. model $M$ could return a claim "Nobuhiro was born in 1976, in Osaka, Japan.", which is still divisible), sampling multiple questions complements the claim extraction process to support fine-grained analysis.

We then obtain the set of answers $\mathcal{A}_q$ for each question $q\in\mathcal{Q}_c$ as
\begin{equation}
    \label{eqn:ans_set}
    \mathcal{A}_{q} = \{a_i\}^{n_a}_{i=1}, \quad \text{where}\;a_i\sim M_{T=t}(q,x),
\end{equation}
and $n_a$ is a hyperparameter specifying the number of answers to generate for each question.

\subsection{Claim-Level Faithfulness}
\label{sec:claim-level_consistency}
Since the answer set obtained in \autoref{eqn:ans_set} is independent of the generative context $R$, it serves as a truthful representation of the model's knowledge and a testing criterion for the information claimed in $R$. Specifically, we acquire the semantic relationship between \autoref{eqn:ans_set} and \autoref{eqn:claims_set} to check if the model has fabricated information. One common approach is to incorporate the Natural Language Inference model to infer the entailment relationship (\citealp{kuhn2023semanticUncertainty}; \citealp{lin2024generatingConfidence}). However, it requires an exhaustive check of pairwise combinations between elements in the answer set and the claim set. Instead, we directly prompt the model $M$ to check if the answer set contradicts the previous claims and output the percentage of contradiction.

Given an atomic claim $c_i$ extracted from $R$, we denote $C_{i\leq}$ as all claims that precede and include $c_i$. We are then interested in knowing whether the model truly possesses the knowledge of $c_i$ or fabricates $c_i$ from the generative context. This is achieved by using the model $M$ to provide an estimate of the percentage of contradiction between $\mathcal{A}_q$ and $C_{i\leq}$, denoted as $X(\mathcal{A}_q,C_{i\leq})$. We define the faithfulness for claim $c_i\in R$ as
\begin{equation}
    \label{eqn:claim_faith}
    F(c_i) = 1-\frac{1}{|\mathcal{\mathcal{Q}}_{c_i}|}\sum_{q\in \mathcal{Q}_{c_i}} X(\mathcal{A}_q,C_{i\leq}),
\end{equation}
where $|\mathcal{Q}_{c_i}|$ is the number of questions in $\mathcal{Q}_{c_i}$. If the claim $c_i$ is a faithful representation of the model's knowledge, $X(\mathcal{A}_q,C_{i\leq})$ will be zero for all questions, and thus $F(c_i)=1$. The average faithfulness for claims in $R$ then constitutes the faithfulness of the response $R$
\begin{equation}
    \label{eqn:res_faith}
    F(R)=\frac{1}{|\mathcal{C}^R|}\sum_{c_i\in \mathcal{C}^R}F(c_i).
\end{equation}
Here, $F(R)$ indicates the tendency of the model to fabricate information, as lower $F(R)$ suggests the possibility that the model's response contradicts its knowledge.

\begin{figure*}[t]
  \centering
  \begin{subfigure}[b]{0.5\linewidth}
    \captionsetup{labelfont=footnotesize}
    \includegraphics[width=\linewidth, height=5.55cm]{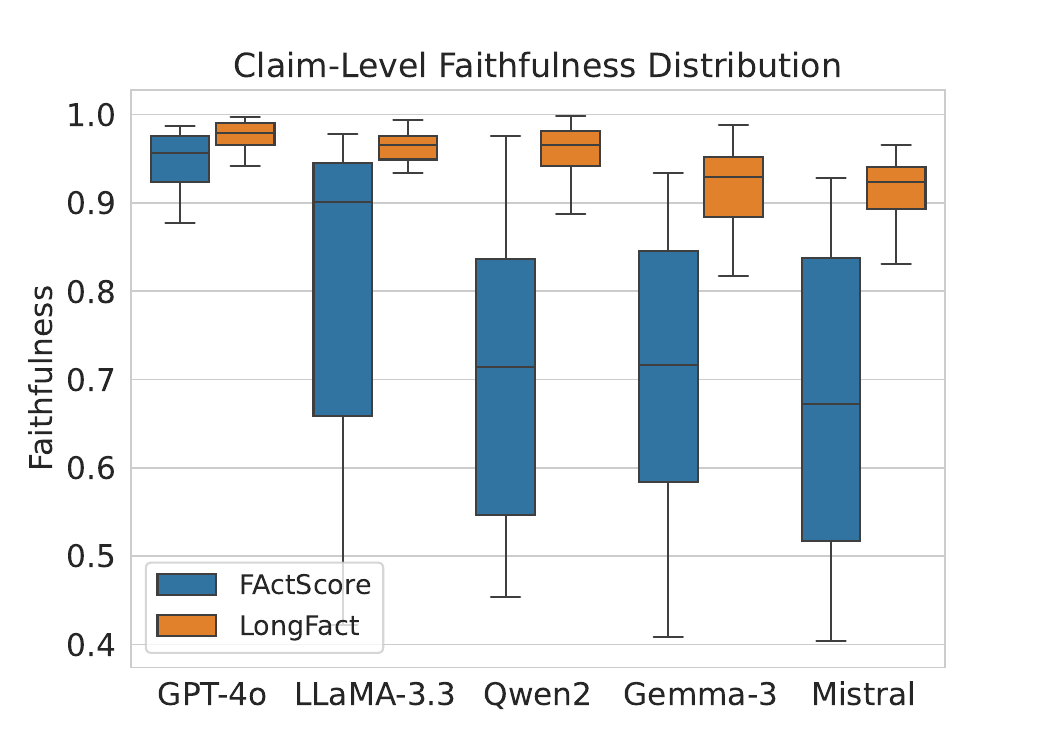}
    \caption{\footnotesize Distribution of all claim faithfulness by models.}
    \label{fig:score_distribution}
  \end{subfigure}
  \hfill
  \begin{subfigure}[b]{0.48\linewidth}
    \captionsetup{labelfont=footnotesize}
    \includegraphics[width=\linewidth, height=5.33cm]{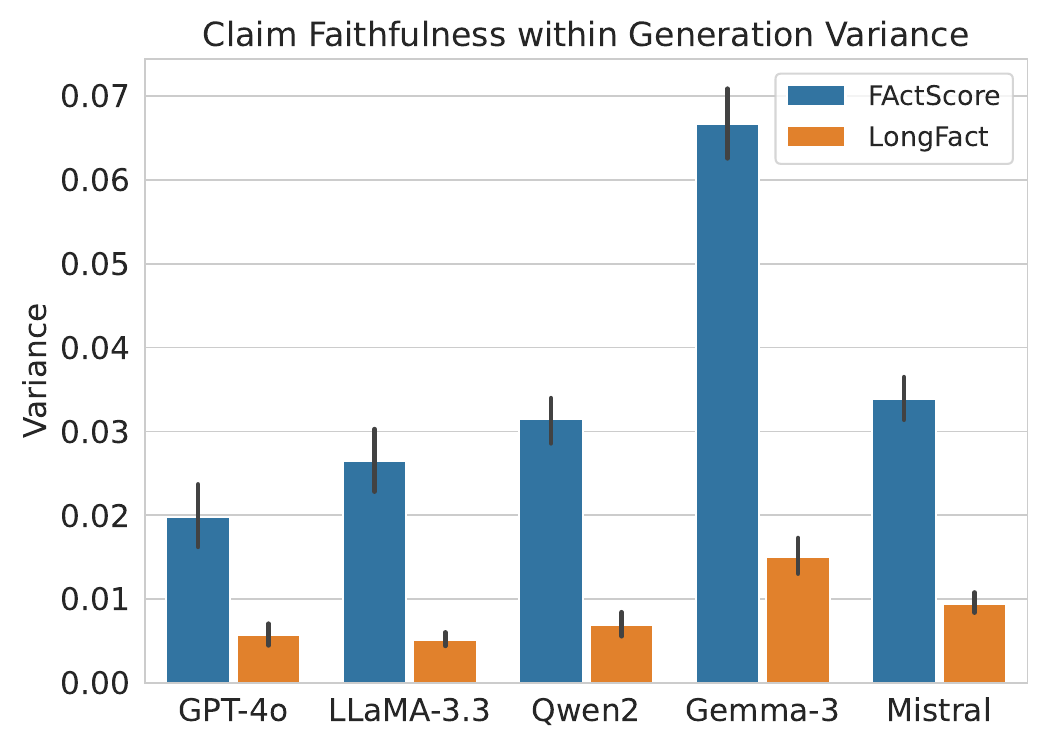}
    \caption{\footnotesize Variance of claim faithfulness within generated responses.}
    \label{fig:variance_within_gen}
  \end{subfigure}
  \caption{Statistics of the claim-level faithfulness over selected models. (a) The faithfulness scores of all claims over FActScore and LongFact. Higher scores indicate less contradiction between claims and model's knowledge. (b) The average variance of claim faithfulness within each sampled response.}
  \label{fig:claim_faith_stat}
\end{figure*}

\subsection{Claim-Level Uncertainty}

We build uncertainty estimation on inter-sample consistency while accounting for the influence of claim faithfulness defined in \autoref{eqn:claim_faith}.
Following \citep{jiang2024graphbased} and \citep{zhang2024luq}, the inter-sample consistency is evaluated at the claim-level by aggregating the number of sampled responses that entail claim $c_i\in\mathcal{C}^R$. For each claim $c_i$, the consistency score is computed as $S(c_i)=\frac{1}{N}\sum_{k=1}^N \mathbbm{1} [R^{k}\Rightarrow c_i]$, where $N$ is the number of sampled responses. However, $S(c_i)$ does not take into account the impact of conditioning on unfaithful information when generating new tokens.

Therefore, we propose to utilize the sequential property of $c_i\in R$ to model the influence of unfaithfulness $1-F(c_i)$. Specifically, we convolve the sequence $[1-F(c_1), 1-F(c_2),\dots,1-F(c_k)]$ with a influence kernel $E$ to propagate the impact of unfaithful claims to subsequent claims. We then define the unfaithfulness weighting for claim $c_i$ as
\begin{equation} 
    W(c_i) = \sum_{j=1}^{i} (1 - F(c_j)) \cdot E(i - j).
    \label{eqn:faith_weight}
\end{equation}
We choose the kernel $E$ as the exponential decay function $E(i)=e^{-\lambda i}$ for claim indices $i=0,1,\dots,k$. Different choices of kernels are evaluted in \autoref{tab:ablation_err_prop}. The uncertainty score for claim $c_i$ is then $S$ scaled by the influence of claim unfaithfulness
\begin{equation}
    U(c_i) = S(c_i) \cdot W(c_i).
\end{equation}

The uncertainty score $U$ encompasses inter-sample consistency while accounting for the influence of preceding unfaithful claims. The weighting $W$ can also be applied to other consistency-based methods to consider the intra-sample claims dependence.

\subsection{Answer-Level Uncertainty}
One major challenge for long-form UQ comes from the large number of generated tokens. Empirically, token-probability based approaches become less effective when the length of generation increases. Leveraging the convenience of short-answers, we reformulate token-based approaches by operating on answer sets $\mathcal{A}_{q}$, where the answers are treated independently as the model's response.

By characterizing the language generation as a classification problem, the uncertainty of an response can be measured by the entropy of the prediction (\citealp{wellmann2012infoEntropyasQualityMeasure}; \citealp{kuhn2023semanticUncertainty}). In general, the predictive entropy (PE) for input x is the conditional entropy ($H$) of the output $R$:
\begin{equation}
    H(R|x)=-\sum_ip(z_i|x)\log{p(z_i|z_{<i},x)},
\end{equation}
\noindent where $z_i$ is the i-th token generated by the LLM and $z_{i<}$ is all the tokens that precedes $z_i$.

As a result, we propose an indirect approach based on token-probability of the answers in the set $\mathcal{A}_q$ without the need to tackle the original long-form response. Since the context is bound to claim $c_i$, their token-probabilities are indicative of the LLM's uncertainty over the claim $c_i$. We define the uncertainty estimate built on entropy $H$ as
\begin{equation}
    U_{A}(c_i) = \frac{1}{|\mathcal{\mathcal{Q}}_{c_i}|}   \sum_{q\in \mathcal{\mathcal{Q}}_{c_i}}  \frac{1}{|\mathcal{A}_q|}  \sum_{a\in\mathcal{A}_Q} H(a|c_i).
    \label{eqn:U_EC}
\end{equation}

The uncertainty quantification result of $U_A$ is shown in \autoref{tab:instruction_tuned_model_results}, complementing the semantic-based approach of IUQ. Methods including perplexity and maximum token entropy can be employed for additional comparison but are excluded for clarity.

\section{LLM Faithfulness}

In this section, we present the analysis of how faithful the model is in generating long-form responses and quantify the model's tendency to fabricate information.

We first collect all claims tested across the datasets and their corresponding faithfulness defined in \autoref{eqn:claim_faith}. The analysis is performed based on: (a) the overall distribution of claim faithfulness across all data samples, and (b) the variance of claim faithfulness within each sampled response. We make the following observation from \autoref{fig:claim_faith_stat}: (i) The average claim faithfulness exhibits a clear distinction among different datasets, as the models are often less faithful with the topics in FActScore than in LongFact. A possible explanation is that FActScore mainly contains biographies of lesser-known individuals, while LongFact addresses popular topics in various fields. (ii) This distinction is also evident in the variance of claim faithfulness within generation, showing that the models are more likely to mix faithful claims with fabricated information.

Given the definition of response faithfulness in \autoref{eqn:res_faith} , the model's faithfulness on the prompt $x$ and the underlying topic is then $F(\mathcal{R}) = \frac{1}{|\mathcal{R}|}\sum_{R\in\mathcal{R}}F(R)$,
where $|\mathcal{R}|$ is the number of sampled claims. We then average $F(\mathcal{R})$ over all topics in the dataset to define the faithfulness of a model as
\begin{equation}
    F(M)=\frac{1}{|D|}\sum_{\mathcal{R}\in D}F(\mathcal{R}).
\end{equation}
Computing $F(M)$ for selected models, we present the quantified model faithfulness in \autoref{tab:model_faith}.
\begin{table}[H]
  \centering
  \renewcommand{\arraystretch}{1.4}
  \footnotesize % Slightly smaller than \small
  \setlength{\tabcolsep}{1pt} % Let \extracolsep handle the spacing
  \begin{tabular*}{\linewidth}{@{\extracolsep{\fill}} l ccccc}
    \toprule
    \textbf{Dataset} & \textbf{GPT} & \textbf{LLaMA3.3} & \textbf{Qwen} & \textbf{Gemma} & \textbf{Mistral} \\
    \midrule
    FActScore & 0.927 & 0.816 & 0.700 & 0.697 & 0.679 \\
    LongFact  & 0.974 & 0.959 & 0.954 & 0.919 & 0.911 \\
    \bottomrule
  \end{tabular*}
  \caption{Model faithfulness on FActScore and LongFact.}
  \label{tab:model_faith}
\end{table}

$F(M)$ serves as a measure of the model's tendency to fabricate information in long-form responses.

\newcommand{\ul}[1]{\underline{#1}}
\newcommand{\bt}[1]{\textbf{#1}}

\begin{table*}[htbp]
    \renewcommand{\arraystretch}{1.3}
    \centering
    \footnotesize
    \adjustbox{width=\textwidth} {
    \begin{tabular}{c|
                    c|
                    >{\centering\arraybackslash}p{1.8cm}
                    >{\centering\arraybackslash}p{1.8cm}
                    >{\centering\arraybackslash}p{1.8cm}
                    >{\centering\arraybackslash}p{1.8cm}
                    >{\centering\arraybackslash}p{1.8cm}
                    >{\centering\arraybackslash}p{1.8cm}
                    |>{\centering\arraybackslash}p{1.8cm}} \toprule
    \multicolumn{1}{c}{} & \multicolumn{1}{|c|}{\textbf{Metric}} & GPT-4o & LlaMA-3.1 & LlaMA-3.3 & Qwen2 & Gemma-3 & Mistral & \multicolumn{1}{|c}{\textbf{Avg.}} \\
    \midrule
    \multirow{9}{*}[1em]{\centering\rotatebox[origin=c]{90}{\textbf{FActScore}}}
        & Max Token Ent. & -          & 0.596      & 0.672      & 0.637      & 0.625      & 0.659  & 0.638 \\
        & PPL            & -          & 0.577      & 0.661      & 0.622      & 0.593      & 0.647  & 0.620\\
        & CCP            & -          & 0.623      & 0.663      & 0.683      & 0.627      & 0.659  & 0.651\\
        & Freq. Scoring  & -          & 0.751      & 0.724      & 0.763      & 0.579      & 0.747  & 0.713\\
        & $U_{A}$        & 0.617      & 0.634      & 0.633      & 0.838      & 0.706      & 0.799  & 0.705\\
        & $S$            & 0.732      & 0.819      & \ul{0.847} & 0.901      & 0.820      & \ul{0.880}  & 0.833\\
        & $C_{C}$        & \bt{0.749}  & \ul{0.822} & 0.843      & \ul{0.929} & \ul{0.840} & 0.862  & \ul{0.841} \\
        & IUQ (Ours)     & \ul{0.748} (\color{amaranth}{-0.1\%})   & \bt{0.847} (\color{ao(english)}{+2.5\%}) & \bt{0.875} (\color{ao(english)}{+2.8\%}) & \bt{0.932} (\color{ao(english)}{+0.3\%}) & \bt{0.867} (\color{ao(english)}{+2.7\%}) & \bt{0.913} (\color{ao(english)}{+3.3\%})  & \bt{0.864} (\color{ao(english)}{+2.3\%}) \\
    \midrule
    \midrule
    \multirow{9}{*}[1em]{\rotatebox[origin=c]{90}{\textbf{LongFact}}} 
        & Max Token Ent. & -           & 0.552      & 0.521      & 0.558      & 0.528      & 0.554  & 0.543\\
        & PPL            & -           & 0.569      & 0.518      & 0.577      & 0.524      & 0.572  & 0.552\\
        & CCP            & -           & 0.559      & 0.520      & 0.508      & 0.537      & 0.539  & 0.533\\
        & Freq. Scoring  & -           & 0.643      & 0.666      & 0.699      & 0.559      & 0.696  & 0.653\\
        & $U_{A}$        & 0.592       & 0.573      & 0.591      & 0.659      & 0.557      & 0.625  & 0.600\\
        & $S$            & 0.705       & \ul{0.736} & \ul{0.714} & \ul{0.791} & \ul{0.656} & \ul{0.733}  & \ul{0.723}\\
        & $C_{C}$        & \ul{0.722}  & 0.724      & 0.702      & 0.782      & 0.639      & 0.712  & 0.714\\
        & IUQ (Ours)     & \bt{0.733} (\color{ao(english)}{+1.1\%})  & \bt{0.749} (\color{ao(english)}{+1.3\%}) & \bt{0.722} (\color{ao(english)}{+0.8\%}) & \bt{0.806} (\color{ao(english)}{+1.5\%}) & \bt{0.689} (\color{ao(english)}{+3.3\%}) & \bt{0.743} (\color{ao(english)}{+1.0\%}) & \bt{0.740} (\color{ao(english)}{+1.7\%}) \\
    \bottomrule
    \end{tabular}

    }
    
    \caption{AUROCs of the uncertainty quantification metrics across models of diverse sizes. Bold-text indicates the highest scores, and italic-text indicates the second highest scores. The White-box uncertainty quantification results for GPT-4o is unavailable due to its closed-source nature.}
    \label{tab:instruction_tuned_model_results} 
    \vspace{-1mm}
\end{table*}

\section{Experiments}
In this section, we present the setup of the experiments and ablation studies to demonstrate the effectiveness of IUQ.

\subsection{Baselines}
\label{sec:baselines}
We select both white-box and black-box uncertainty quantification methods as baselines, using implementations from \citep{fadeeva2023lm-polygraph} for white-box methods. Specifically, we include:
\begin{itemize}[leftmargin=*]
    \item \textbf{Max Token Entropy} \citep{fomicheva2020maxTokenEntropy}: This method quantifies uncertainty by calculating the Shannon entropy of the generated tokens. For a given output sequence, it identifies the maximum entropy value across the aligned tokens of claims in the original response, where high entropy at any step indicates a lack of confidence in the token selection.

    \item \textbf{Perplexity (PPL)} \citep{fomicheva2020maxTokenEntropy}: Perplexity represents the geometric mean of the inverse probability of the tokens. For claim-level analysis, the extracted claims are aligned with the original response to use the corresponding tokens.

    \item \textbf{Claim-Conditioned Probability (CCP)} \citep{fadeeva2024ccp}: This method improves upon the entropy-based method by isolating factual uncertainty from linguistic variation. It identifies the semantically important tokens and takes into account the probabilities of their alternatives. This strategy effectively leverages information encapsulated in the output without the need to perform additional sampling.

    \item \textbf{Frequency Scoring} \citep{mohri2024freqScoring}: This approach samples multiple alternative responses from the model to define the associated uncertainty sets, where each set contains statements that entail the model’s output. The author show how conformal prediction defines a back-off algorithm for ensuring the correctness of LM outputs, and correspondingly define a uncertainty metric at claim-level.

    \item \textbf{Claim Entailment ($S$)}: This approach is adopted in \citep{zhang2024luq} and \citep{jiang2024graphbased} to evaluate the uncertainty of a sequence by aggregating the number of sampled generations that entail the sequence. For claim-level analysis, the score $S$ for claim $c_i$ is computed as $S(c_i)=\frac{1}{N}\sum_{k=1}^N \mathbbm{1} [R^{k}\Rightarrow c_i]$, where $N$ is the total number of sampled responses.

    \item \textbf{Closeness Centrality ($C_C$)} \citep{jiang2024graphbased}: Building on the Claim Entailment $S$, Closeness Centrality is a graph-based method that exploits the connectivity of nodes to estimate the likelihood for the claim to hold true. A bipartite graph is constructed by treating each claim as a node and drawing an edge between the node and sampled responses according to their entailment relationship. While multiple graph-based methods are explored (betweenness, eigenvalue, PageRank), we only compare Closeness Centrality ($C_C$), which is the best-performing metric.
\end{itemize}

\subsection{Datasets and Annotation}
\label{sec:datasets_and_annotation}

\begin{table}[ht]
  \centering
  \small
  \renewcommand{\arraystretch}{1.4}
  \begin{tabularx}{\linewidth}{l>{\centering\arraybackslash}X>{\centering\arraybackslash}X>{\centering\arraybackslash}X>{\centering\arraybackslash}X}
    \toprule
    \textbf{Dataset} & \textbf{Responses} & \textbf{Claims} & \textbf{Questions} & \textbf{Answers} \\
    \hline
    FActScore & 235 & 4759 & 10433 & 31299 \\
    LongFact  & 250 & 4276 & 9954  & 29862 \\
    \textbf{Total} & \textbf{485} & \textbf{9035} & \textbf{20387} & \textbf{61161} \\
    \bottomrule
  \end{tabularx}
  \caption{Statistics of the total numbers of generated items by GPT-4o on FActScore and LongFact.}
  \label{tab:gpt-4o_data_stat}
\end{table}

We evaluate IUQ on FActScore \citep{min2023factscore} and LongFact\citep{wei2024longfact}. We select entities from each dataset, using the provided prompt as input and the reference text to check for claim-level correctness. A statistics of the data composition is shown in \autoref{tab:gpt-4o_data_stat}. The processing of each dataset is as follows:

\begin{table*}[htbp]
\renewcommand{\arraystretch}{1.3}
\centering
\footnotesize
\setlength{\tabcolsep}{2.5pt} % Default is usually 6pt
\adjustbox{width=\textwidth} {
    \begin{tabular}{c
                    >{\centering\arraybackslash}p{1.65cm}
                    >{\centering\arraybackslash}p{1.65cm}
                    >{\centering\arraybackslash}p{1.65cm}
                    >{\centering\arraybackslash}p{1.65cm}
                    >{\centering\arraybackslash}p{1cm}
                    >{\centering\arraybackslash}p{1.65cm}
                    >{\centering\arraybackslash}p{1.65cm}
                    >{\centering\arraybackslash}p{1.65cm}
                    >{\centering\arraybackslash}p{1.65cm}
                    >{\centering\arraybackslash}p{1cm}}
    \toprule
    \multirow{2}{*}{\textbf{Method}} & \multicolumn{5}{c}{\textbf{FActScore}} & \multicolumn{5}{c}{\textbf{LongFact}} \\
    \cmidrule(lr){2-6} \cmidrule(lr){7-11}
    & \textbf{GPT-4o} & \textbf{LlaMA-3.3} & \textbf{Qwen2} & \textbf{Gemma-3} & \textbf{Avg.} & \textbf{GPT-4o} & \textbf{LlaMA-3.3} & \textbf{Qwen2} & \textbf{Gemma-3} & \textbf{Avg.} \\
    \midrule
        Lin-E   & \ul{0.732} & 0.858 & 0.917  & 0.834 & 0.835 & \ul{0.725} & \ul{0.714} & \ul{0.801} & \ul{0.682} & 0.731 \\
        Acc-E   & 0.713 & 0.841 & 0.889  & 0.804 & 0.812 & 0.723 & 0.710 & 0.800 & 0.675 & 0.727 \\
        No-E    & \bt{0.748} & \ul{0.871} & \ul{0.931}  & \ul{0.847} & \ul{0.849} & 0.724 & \bt{0.722} & \bt{0.806} & 0.678 & \ul{0.733} \\
        Exp-E (IUQ)   & \bt{0.748} & \bt{0.875} & \bt{0.932} & \bt{0.867} & \bt{0.856} & \bt{0.733} & \bt{0.722} & \bt{0.806} & \bt{0.689} & \bt{0.738} \\
    \bottomrule
    \end{tabular}
}
\caption{Ablation study on the impact of claim consistency score with different error propagation (E) function. The presented values are AUROCs of the uncertainty quantification metric $U_{S}$.}
\label{tab:ablation_err_prop}
\vspace{-1mm}
\end{table*}

\noindent\textbf{FActScore} \citep{min2023factscore} contains entities of human biography, where each of them has a dedicated Wikipedia article. We randomly select 50 entities. To evaluate the factuality of claims, IUQ employs a similar method in \citet{min2023factscore}, labeling each fact as "correct" or "incorrect" based on the corresponding Wikipedia article. The factuality evaluation is independent of the uncertainty estimation process and is performed using GPT-4o.

\noindent\textbf{LongFact} \citep{wei2024longfact} is a prompt set comprising thousands of questions spanning 38 topics. We choose LongFact to test our uncertainty metrics since it complement FActScore on the domains of topics. While FActScore verifies the correctness of atomic claims through reference passages from Wikipedia, the approach proposed in \citep{wei2024longfact} does so by using search engine to fetch and evaluate internet-based sources. To maintain consistency and reproducibility, we manually select 50 entities of diverse topics in LongFact that include dedicated Wikipedia articles, and employ the same method we used for FActScore to evaluate the factuality of claims.

\subsection{Models and Parameters} We conduct experiments over models across various model families, including GPT4o \citep{openai2024gpt4o}, LlaMA-3.1-70B-Instruct, LlaMA-3.3-70B-Instruct \citep{touvron2023llama}, Qwen2-VL-72B \cite{yang2024qwen2}, Gemma-3-27b-it \citep{gemmateam2025gemma3}, and Mistral-Small-24B-Instruct \citep{jiang2023mistral7b}. For each data entity, we sample 5 long-form responses using temperate $t=1.0$, and use temperate $t=0$ to evaluate the correctness of claims.

\subsection{Evaluation Metrics}
Following prior works (\citealp{manakul2023selfcheckgpt}; \citep{kuhn2023semanticUncertainty}; \citealp{jiang2024graphbased}), we formulate the evaluation process as a classification problem, where the predicted  probability of claims being correct is given by our uncertainty metrics, and the procedure to obtain ground-truth labels is detailed in \autoref{appendix:corr_eval}. We adopt the area under the receiver operator characteristic curve (AUROC) and Area Under the Precision-Recall Curve (AUPRC) to classify the performance of the uncertainty metrics.
Additionally, we report the Pearson correlation coefficient between confidence scores and ground-truth correctness labels, along with 95\% confidence intervals and $p$-values, in \autoref{appendix:pearson}.

\subsection{Ablation Study}
In this section, we present an experimental study to show the effectiveness of our claim-consistency paradigm (\autoref{sec:claim-level_consistency}). Firstly, we illustrate that IUQ captures the model's self-contradictory behavior in its response, by comparing the performance of baselines and IUQ metrics. Secondly, by evaluating the influence of using different error propagation functions, we show that the exponential-decay weighting is the most effective approach to estimate uncertainty in long-form generations. Lastly, we evaluate the sensitivity of our uncertainty metrics on the number of generated responses. We present ablation results on selected models in \autoref{tab:ablation_err_prop} and \autoref{fig:ablation_gen_num}. Additional experiments are reported in \autoref{appendix:iuq_rev}.

\noindent\textbf{Effectiveness of Claim Consistency Score}
The claim faithfulness score (\autoref{eqn:claim_faith}) captures the fabricated information in long-form responses by enforcing a consistency check between claims and context. To demonstrate its effectiveness, we compare its performance with verbal-confidence, which is the confidence score elicited from the model. 

The result illustrates that although $S$ is not a particualrly strong baseline, IUQ shows superior performance over all tested models. This observation consolidates our motivation that LLM has limitations in identifying its own lack of knowledge. Without sampling multiple responses and performing fine-grained analysis, it is risky to trust LLM responses, especially in long-form generation.

\noindent\textbf{Effectiveness of Influence Kernel} 
The influence kernel $E$ serves to propagate the impact of an inconsistent claim to subsequent claims. In this section, we investigate the influence of different kernels as propagation functions on the uncertainty estimation performance. The results are shown in \autoref{tab:ablation_err_prop} and the notations used are explained as follows: (1) No-E: No error is propagated to subsequent claims, and we build the uncertainty estimate solely on the claim consistency score. (2) Lin-E: Linear error propagation, where the unfaithfulness scores is superimposed with a linear function $f(k)=mi+b$ for $i=k,k-1,\dots, 1$, where $m>0$ and $b$ is a constant. (3) Acc-E: accumulative error propagation, where the cumulative sums of the claim-level unfaithfulness are used as the weighting to the entailment score $S$.

\noindent\textbf{Influence of Number of Generations}
We show the influence of the number of sampled responses on our uncertainty metric IUQ and black-box baseline methods in \autoref{fig:ablation_gen_num}. The experiments are performed using GPT-4o. The result demonstrates that the number of sampled responses has a non-negligible effect in quantifying the uncertainty, as more samples leads to more accurate claim-entailment scrores $S$.

\begin{figure}[t]
  \includegraphics[width=\columnwidth]{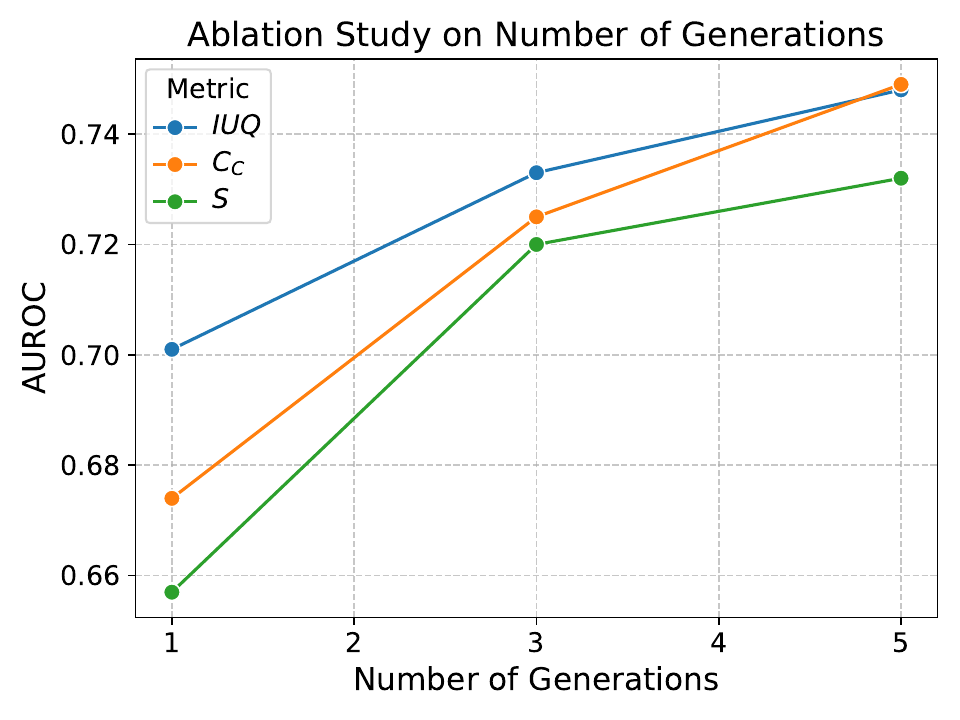}
  \caption{AUROCs of IUQ and baselines on different numbers of sampled responses.}
  \label{fig:ablation_gen_num}
\end{figure}

\section{Conclusion}
We propose Interrogative Uncertainty Quantification (IUQ) that builds on claim-level contextual consistency to estimate the uncertainty in long-form responses. Empirical results demonstrate the effectiveness of IUQ over diverse model families.

\section{Limitations}
The limitations of our study include the following:
\textbf{(1)} Our method relies on LLMs' reasoning and question-answering ability to perform most parts of our pipeline. A major issue is the possible hallucination introduced in the workflow, and there is no guarantee that such hallucination will be detected. This problem is partially addressed by adapting the source code to incorporate the model provider's support of structured output, which is limited to a few latest models. Additional measures we take are to manually parse the model's output and perform heuristic sanity checks to ensure model responses are at least sensible. \textbf{(2)} In this study, we do not process the model response that refuse to answer, for example, when it declare a lack of knowledge on the specific topic. Our consideration is to ensure a fair comparison between our method and the baselines, and avoid injecting bias by manually processing the event of response refusal.
\textbf{(3)} IUQ pipeline requires multiple stages of LLM inference, and thus may incur unexpectedly large computation cost. Specifically, IUQ requires approximately 47.3\% more tokens than Graph-Based Uncertainty \citep{jiang2024graphbased}; we provide a stage-by-stage breakdown in~\autoref{appendix:compute_cost}.

\section{Ethics Statement}
This work includes individuals whose information is used solely for research purposes. No attempt is made to acquire information from sources other than publicly available websites and datasets. This work does not involve personally identifiable data or sensitive information. All experiments are conducted using publicly available data and therefore raise no direct ethical or privacy concerns. We have carefully adhered to the ACL Guidelines of Ethics throughout the research and writing process.

\section{AI Assistance Statement}
The use of AI tools was solely to assist with the linguistic polishing of this manuscript, such as improving grammar, clarity, and readability. All conceptual contributions, technical methods, experimental designs, and analyses were developed entirely by the authors without the use of AI.


\bibliography{custom}

\clearpage

\appendix

\section*{Appendix}
\section{Validation of Contradiction Evaluation: IUQ-rev}
\label{appendix:iuq_rev}

While the Claim Entailment baseline $S$ already establishes a good uncertainty estimate from cross-sample inconsistency (see \autoref{tab:instruction_tuned_model_results}), our hypothesis that model unfaithfulness conditioned on prior context is not tested. Therefore, we present an ablation study to probe a critical question: does the \emph{direction} of the contradiction check matter?

Specifically, IUQ evaluates contradiction against preceding claims $C_{i\leq}$, grounded in the hypothesis that fabrication is driven by conditioning on prior context. \textbf{IUQ-rev} is a controlled variant that instead uses subsequent claims $C_{i>}$: since subsequent claims are the causally downstream of claim $c_i$, they carry no information about whether $c_i$ was fabricated to fit prior context. If the gain of IUQ were a generic artifact of any contradiction signal, IUQ-rev would perform comparably.

\begin{table}[H]
  \centering
  \small
  \renewcommand{\arraystretch}{1.4}
  \setlength{\tabcolsep}{5.5pt}
  \begin{tabular}{l
                  >{\centering\arraybackslash}p{1.2cm}
                  >{\centering\arraybackslash}p{1.2cm}
                  >{\centering\arraybackslash}p{2cm}}
    \toprule
    \textbf{} & \textbf{Ent. ($S$)} & \textbf{IUQ} & \textbf{IUQ-rev} \\
    \midrule
    \multicolumn{4}{l}{\textit{\textbf{FActScore}}} \\
    GPT-4o     & $0.732$ & 0.748 & 0.738\,(\color{amaranth}{$-$\%1.0}) \\
    LLaMA-3.1  & $0.819$ & 0.847 & 0.832\,(\color{amaranth}{$-$\%1.5}) \\
    LLaMA-3.3  & $0.847$ & 0.875 & 0.863\,(\color{amaranth}{$-$\%1.2}) \\
    Qwen2      & $0.901$ & 0.932 & 0.916\,(\color{amaranth}{$-$\%1.6}) \\
    Gemma-3    & $0.820$ & 0.867 & 0.832\,(\color{amaranth}{$-$\%3.5}) \\
    Mistral    & $0.880$ & 0.913 & 0.896\,(\color{amaranth}{$-$\%1.7}) \\
    \midrule
    \multicolumn{4}{l}{\textbf{\textit{LongFact}}} \\
    GPT-4o     & $0.705$ & 0.733 & 0.716\,(\color{amaranth}{$-$\%1.7}) \\
    LLaMA-3.1  & $0.736$ & 0.749 & 0.737\,(\color{amaranth}{$-$\%1.2}) \\
    LLaMA-3.3  & $0.714$ & 0.722 & 0.724\,(\color{ao(english)}{$+$\%0.2}) \\
    Qwen2      & $0.791$ & 0.806 & 0.791\,(\color{amaranth}{$-$\%1.5}) \\
    Gemma-3    & $0.656$ & 0.689 & 0.681\,(\color{amaranth}{$-$\%0.8}) \\
    Mistral    & $0.733$ & 0.743 & 0.735\,(\color{amaranth}{$-$\%0.8}) \\
    \bottomrule
  \end{tabular}
  \caption{
    AUROC comparison between Claim Entailment score ($S$), IUQ and IUQ-rev. IUQ-rev replaces the preceding-claims context $C_{i\leq}$ with subsequent claims $C_{i>}$ in the Contradiction Evaluation stage. Percentages in parentheses are relative to IUQ.
  }
  \label{tab:iuq_rev}
\end{table}

As shown in \autoref{tab:iuq_rev}, IUQ-rev underperforms IUQ on all models for FActScore and on most models for LongFact, confirming that the directional design is essential and validating the contextual-dependence hypothesis.

\section{Statistical Significance: Pearson Correlation}
\label{appendix:pearson}

Considering our experiments are conducted on a small portion (50 samples each) of the FActScore and LongFact datasets, we demonstrate the statistical significance by reporting the Pearson correlation coefficient $r$ between the binary claim correctness labels ($\in\{0,1\}$) and the negative uncertainty scores (confidence scores) for the three best-performing methods. The result is shown in~\autoref{tab:pearson_correlation}. For clarity, we only report the results for Closeness Centrality ($C_C$)~\citep{jiang2024graphbased} and IUQ in~\autoref{tab:pearson_correlation}.

\begin{table}[H]
  \centering
  \small
  \renewcommand{\arraystretch}{1.4}
  \setlength{\tabcolsep}{5pt}
  \begin{tabular}{lcc}
    \toprule
    \textbf{} & \textbf{Close Cent. ($C_C$)} & \textbf{IUQ (Ours)} \\
    \midrule
    \multicolumn{3}{l}{\textit{\textbf{FActScore}}} \\
    GPT-4o     & 0.377\,\ci{0.353}{0.401} & 0.378\,\ci{0.353}{0.402} \\
    LLaMA-3.1  & 0.507\,\ci{0.479}{0.534} & 0.548\,\ci{0.522}{0.574} \\
    LLaMA-3.3  & 0.575\,\ci{0.554}{0.596} & 0.636\,\ci{0.617}{0.654} \\
    Qwen2      & 0.669\,\ci{0.646}{0.691} & 0.741\,\ci{0.722}{0.759} \\
    Gemma-3    & 0.542\,\ci{0.524}{0.560} & 0.624\,\ci{0.608}{0.639} \\
    Mistral    & 0.543\,\ci{0.519}{0.567} & 0.703\,\ci{0.686}{0.720} \\
    \midrule
    \multicolumn{3}{l}{\textit{\textbf{LongFact}}} \\
    GPT-4o     & 0.259\,\ci{0.231}{0.287} & 0.298\,\ci{0.271}{0.326} \\
    LLaMA-3.1  & 0.298\,\ci{0.271}{0.325} & 0.359\,\ci{0.333}{0.385} \\
    LLaMA-3.3  & 0.276\,\ci{0.248}{0.303} & 0.341\,\ci{0.314}{0.367} \\
    Qwen2      & 0.342\,\ci{0.300}{0.382} & 0.418\,\ci{0.379}{0.455} \\
    Gemma-3    & 0.181\,\ci{0.154}{0.207} & 0.264\,\ci{0.238}{0.289} \\
    Mistral    & 0.257\,\ci{0.224}{0.289} & 0.302\,\ci{0.270}{0.333} \\
    \bottomrule
  \end{tabular}
  \caption{
    Pearson $r$ with 95\% \textcolor{black!55}{confidence-interval} between the binary claim correctness label and the negative uncertainty (confidence) on \textbf{FActScore} and \textbf{LongFact}.
  }
  \label{tab:pearson_correlation}
\end{table}

Across both benchmarks, IUQ generally shows higher Pearson'r, indicating a stronger positive association between confidence scores and claim correctness. The 95\% confidence intervals are computed using Fisher's $z$-transformation and all associated $p$-values are $p < 0.001$.

\section{Computational Cost Analysis}
\label{appendix:compute_cost}

\begin{table*}[htbp]
  \centering
  \small
  \renewcommand{\arraystretch}{1.35}
  \adjustbox{width=\textwidth}{
  \begin{tabular}{lcccccccc}
    \toprule
    & \textbf{Greedy} & \textbf{Diverse} & \textbf{Claim} & \textbf{Claim} & \textbf{Claim} & \textbf{Question} & \textbf{Answer} & \textbf{Contradiction} \\
    & \textbf{Gen.} & \textbf{Gen.} & \textbf{Extraction} & \textbf{Correctness} & \textbf{Supportness} & \textbf{Gen.} & \textbf{Gen.} & \textbf{Evaluation} \\
    \midrule
    Avg.\ Tokens & 262 & 1,331 & 2,906 & 99,586 & 135,597 & 10,356 & 29,902 & 73,609 \\
    \midrule
    Max Token Ent.        & \checkmark & ---                & \checkmark & \checkmark & ---        & --- & --- & --- \\
    PPL                   & \checkmark & ---                & \checkmark & \checkmark & ---        & --- & --- & --- \\
    CCP                   & \checkmark$^{\dagger}$ & ---   & \checkmark & \checkmark & ---        & --- & --- & --- \\
    Freq.\ Scoring        & ---        & \checkmark$^{\dagger}$ & \checkmark & \checkmark & ---   & --- & --- & --- \\
    Claim Entail.\ ($S$)  & ---        & \checkmark         & \checkmark & \checkmark & \checkmark & --- & --- & --- \\
    Closeness Cent.\ ($C_C$) & \checkmark & \checkmark      & \checkmark & \checkmark & \checkmark & --- & --- & --- \\
    \textbf{IUQ (Ours)}   & ---        & \checkmark         & \checkmark & \checkmark & \checkmark & \checkmark\,(+4.3\%) & \checkmark\,(+12\%) & \checkmark\,(+31\%) \\
    \bottomrule
  \end{tabular}
  }
  \caption{
    Per-stage token consumption and stage requirements for each method.
    \checkmark\ indicates a required stage; $\dagger$ indicates additional use of an NLI model.
    Percentages in the IUQ row denote the incremental token cost relative to the shared-stage total.
  }
  \label{tab:compute_cost}
\end{table*}

We present a breakdown of the average token consumption per inference stage for the baselines and our methods. Conveniently, multiple stages are common among claim-level UQ methods (e.g. response sampling, claim extraction, and claim correctness evaluation.). We compute the sum of prompt and completion tokens then average across all data samples used in our experiments. The percentage increases shown for IUQ's three additional stages are computed relative to the cumulative token cost of the preceding shared stages (Greedy Gen.\ + Diverse Gen.\ + Claim Extraction + Claim Correctness + Claim Supportness = 239{,}682 tokens on average).
Note that the Claim Correctness stage can be omitted at test time when no reference text is available.
Overall, IUQ incurs approximately 47.3\% more token consumption compared to Graph-Based Uncertainty \citep{jiang2024graphbased}.

\begin{figure*}[t]
  \includegraphics[scale=0.42]{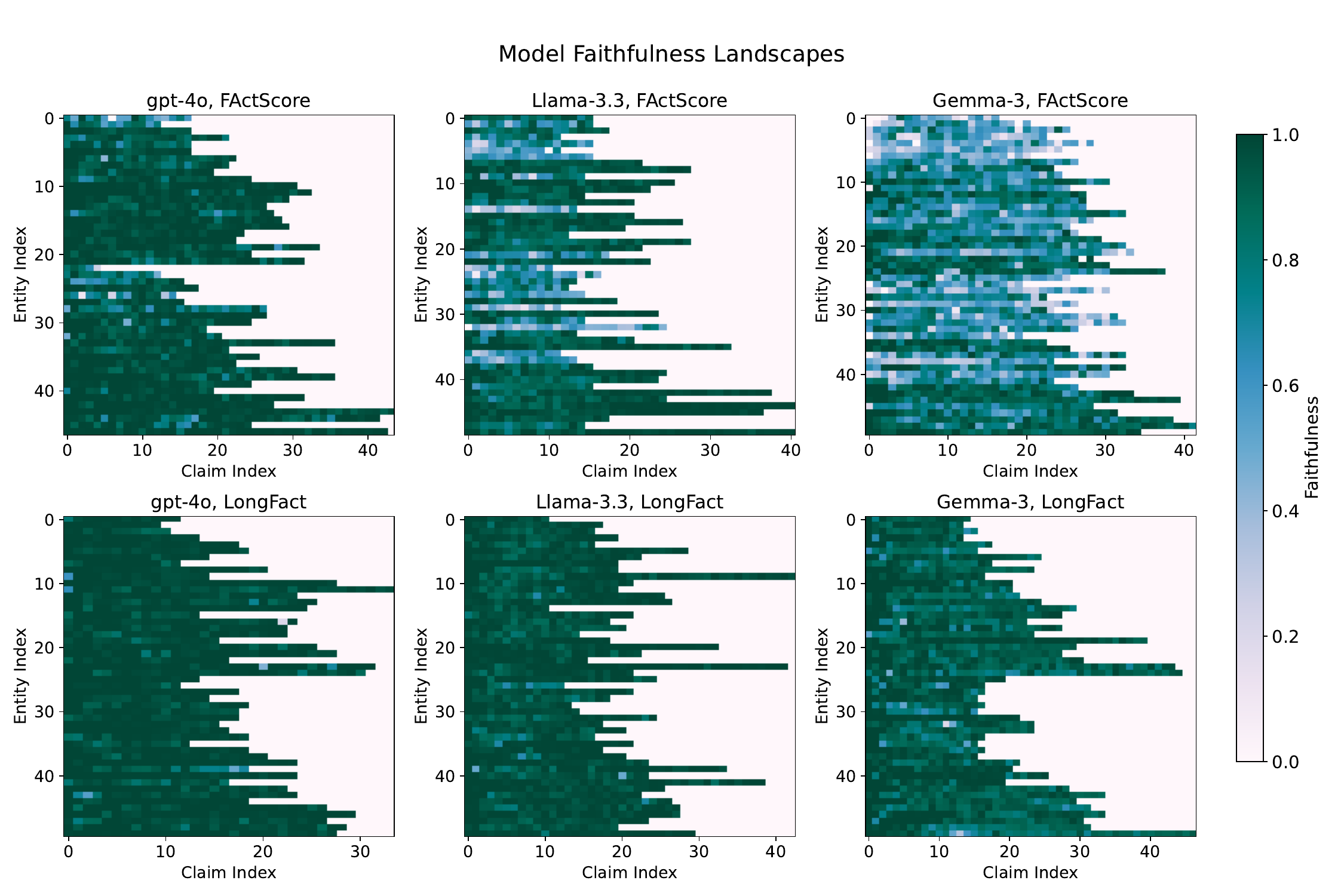}
  \caption{Model faithfulness on claims within individual generation. Results for FActScore and LongFact are shown with selected models. The x-axis is the index of the claim made in LLM's response, and y-axis is the index of the topic in datasets.}
  \label{fig:claim_consistency_landscapes}
\end{figure*}

\section{Case Study: Anecdotal Analysis}
\label{appendix:case_study}

To illustrate how IUQ identifies hallucinations that inter-sample supportness alone cannot detect with an example, we present a claim-level analysis in~\autoref{tab:case_study_rory}. \textit{Rory Byrne} is an entity included in FActScore with moderate exposure on Wikipedia, which is FActScore's ground-truth reference source. According to his Wikipedia page, Rory Byrne is an engineer and car designer~\citep{WikiRoryByrne}.

We use Llama-3.3-70B-Instruct to generate 5 responses, and follow the pipeline of IUQ to obtain the supportness score $S$ and claim faithfulness $F$. The result in~\autoref{tab:case_study_rory} shows the full claim-level breakdown for one of the LLM generation about Rory Byrne. The LLM fabricated a coherent but entirely fictional architect biography, including a firm (``Baxter Byrne Associates''), a landmark project (``Arkangel Tower''), and two named awards — none of which correspond to verifiable facts. All 11 claims are incorrect.

\begin{table}[htbp]
  \centering
  \renewcommand{\arraystretch}{1.5}
  \setlength{\tabcolsep}{3pt}
  \adjustbox{width=\linewidth}{
  \begin{tabular}{c l c c c}
    \toprule
    $c_i$ & \textbf{Claim (abbreviated)} & \textbf{Correct} & $\textbf{\textit{S}}$ & $\textbf{\textit{F}}$ \\
    \midrule

    \rowcolor{yellow!30}
    $c_1$ & ``is an Irish architect'' & $\times$ & \textbf{0.60} & \textcolor{amaranth}{\textbf{0.17}} \\

    \multicolumn{5}{p{0.96\linewidth}}{%
      \vspace{1pt}
      \hspace{1.5em}
      \begin{minipage}{0.93\linewidth}
      \textbf{Question:} \emph{What is Rory Byrne's profession?}

      \begin{tabular}{@{}ll@{}}
      Answer 1: & He is an architect (contr.=0.5) \\
      Answer 2: & He is a civil engineer (contr.=1.0)\\
      Answer 3: & He is a civil engineer (contr.=1.0) \\
      \end{tabular}
      \end{minipage}
      \vspace{4pt}
    } \\

    $c_2$    & ``pioneering in green building \& urban design''    & $\times$ & 0.40          & 0.40 \\
    $c_3$    & ``born in Dublin, Ireland, in 1950''                & $\times$ & 0.40          & 0.58 \\
    $c_4$    & ``founder of Baxter Byrne Associates''              & $\times$ & 0.20          & 0.30 \\
    $c_5$    & ``Baxter Byrne: sustainable \& energy-efficient''   & $\times$ & 0.20          & 0.43 \\
    $c_6$    & ``Arkangel Tower: BIL DEC certified''               & $\times$ & 0.20          & 0.42 \\
    $c_7$    & ``Dublin Docklands Masterplan involvement''         & $\times$ & 0.20          & 0.37 \\
    $c_8$    & ``strong advocate for sustainability in design''    & $\times$ & 0.60 & 0.37 \\
    \rowcolor{green!30}
    $c_9$    & ``President's Award of Excellence in Architecture'' & $\times$ & \textcolor{amaranth}{\textbf{0.20}} & \textbf{0.72} \\

    \multicolumn{5}{p{0.96\linewidth}}{%
      \vspace{1pt}
      \hspace{1.5em}
      \begin{minipage}{0.93\linewidth}
      \textbf{Question:} \emph{What award has Rory received?}
      \begin{tabular}{@{}ll@{}}
      Answer 1: & He received the Royal Gold Medal in architecture (contr.=0.2) \\
      Answer 2: & He received the RIAI Gold Medal in architecture. (contr.=0.4)\\
      Answer 3: & He received the Royal Gold Medal in architecture (contr.=0.25) \\
      \end{tabular}

      \end{minipage}
      \vspace{4pt}
    } \\

    $c_{10}$ & ``Stipan T. Sidoti Award for Urban Design''         & $\times$ & 0.20 & 0.47 \\    
    $c_{11}$ & ``influenced architectural practices globally''     & $\times$ & 0.60 & 0.30 \\
    \bottomrule
  \end{tabular}
  }
  \caption{
    Claim-level analysis of one generation about \textit{Rory Byrne}. All 11 claims are factually incorrect when verified according to his Wikipedia page. \colorbox{yellow!30}{Higher supportness score $S$} indicates the claim is supported by multiple generations, and \colorbox{green!30}{higher faithfulness score $F$} suggests the answers to the corresponding question are consistent with the previous context.
  }
  \label{tab:case_study_rory}
\end{table}

A critical limitation of supportness $S$ is apparent here. Using claim $c_1$ as an example, since three out of five generations are conditioned on the same fabricated premise (Rory Byrne is an Irish architect), they tell similar false stories and mutually corroborate each other, thus obtaining $S=0.60$ for claim $c_1$. The model then needs to generate 3 answers for the question "What is Rory Byrne’s profession". Among these answers, answer 1 is determined by the model to be partially contradictory (contr.=0.5) to $c_1$ while the others are direct contradictions (contr.=1.0). Therefore, the faithfulness score of $c_1$ is $1-(0.5+1.0+1.0\approx 0.17)$. Since $F$ significantly down-weights $S$, the uncertainty assigned to claim $c_1$ is higher than $S$ alone would have suggested.

The exponential-decay influence kernel further amplifies this signal at the generation level.
The early, highly unfaithful claim $c_1$ propagates its penalty to all subsequent claims through $W(c_i)$, reflecting the causal structure of context-conditioned fabrication: once a false premise is established in position $c_1$, every downstream claim is generated under its influence.

For claim $c_9$, the relatively high faithfulness scores suggest that they do not contradict previous context. Given the question, “What award has Rory received?”, mentioning different awards is not necessarily inconsistent with the context, even if those awards are fabricated. However, the low $S$ score for $c_9$ indicates that it is not corroborated by other generations. This suggests that claim-level uncertainty has two distinct aspects: dependence on the prior context and consistency across generations.

\section{Visualization}
\label{appendix:visualization}

\begin{table*}[htbp]
  \centering
  \adjustbox{width=\textwidth} {
  \begin{tabular}{|c|c|}
    \hline
    \textbf{LongFact Prompt} & \textbf{Wiki-entry}  \\
    \hline
    Can you describe the occurrences during the Watts Riots? & Watts riots \\
    \hline
    Can you provide an overview of the International Monetary Fund? & International Monetary Fund \\
    \hline
    Could you explain what the Kepler Space Telescope is? & Kepler space telescope \\
    \hline
  \end{tabular}
  }
  \caption{Example LongFact prompts and corresponding Wikipedia entries.}
  \label{tab:longfact_prompts}
\end{table*}

\noindent\textbf{Model Faithfulness Landscapes}
The faithfulness weighting computed in \autoref{eqn:faith_weight} encapsulates the faithfulness of the claim within LLM generation. Since every claim is assigned a weighting, we can visualize all scores for an entire run of experiment, as shown in  \autoref{fig:claim_consistency_landscapes}. 

To accommodate multiple samples of response, each having different numbers of claims, we interpolate the faithfulness weighting of shorter responses linearly to obtain sets with equal numbers of elements. The sequence of faithfulness weighting representing a single topic is then averaged across the interpolated sequences.

The visualization is in accordance with the claim-level faithfulness distribution shown in~\autoref{fig:claim_faith_stat}, where the models generally exhibit good consistency with LongFact, which mainly contains well-known topics, and poorer consistency with FActScore, which contains less famous individuals. It can also be observed that larger and stronger models are more consistent and thus more faithful in their answers.

\section{Claim-Level Correctness}
\label{appendix:corr_eval}

\textbf{FActScore} We evalute the factual correctness of claims extracted from long-form responses using an adapted approach in \citet{min2023factscore}. For each topic, first, the reference article is fetched from Wikipedia and broken into chunks of passages. The passages and claims are vectorized using sentence-transformer gtr-t5-large \citet{ni2021large-gtr}. Based on the relevance of the claim and the reference passage, the passages are returned based on similarity. The correctness of claims are evaluated by GPT-4o and labeled as either "correct" or "incorrect".

\noindent\textbf{LongFact} LongFact is a dataset that contains 2,280 prompts that solicit long-form responses across 38 selected topics, including arts, chemistry, historical events and etc. \citet{wei2024longfact} propose to use Google Search API to exhaustively verify the factuality for each fact presented in the long-form response. However, to maintain consistency and reproducibility, we manually selected 50 prompts from LongFact that have dedicated Wikipedia entries, and use the same method for FActScore to evaluate factual correctness. Example prompts and Wikipedia entities for LongFact are shown in~\autoref{tab:longfact_prompts}.

\section{Prompts}
\label{appendix:prompts}
We follow the structure of \autoref{fig:pipeline} to list the prompts used in IUQ~\autoref{tab:pipeline_prompts}. Generally, they include the prompts used on generating long-form responses, performing claim-level question answering, and evaluating consistency.

\clearpage
\onecolumn

\begin{table*}[t]
  \renewcommand{\arraystretch}{1.5}
  \centering
  \footnotesize
  \adjustbox{width=\textwidth}{
  \begin{tabular}{|p{9cm}|p{3cm}|}
    \hline
    \multicolumn{1}{|c}{\textbf{Prompt}} & \multicolumn{1}{|c|}{\textbf{Role}} \\
    \hline
    "Answer the following question in plain text, without any additional formatting: \{prompt\}" & Generate response \\
    \hline
    "Given context and a paragraph of text, deconstruct the text into the smallest possible standalone and self-contained facts without semantic repetition. Each fact should come from the text and must be related to the context.\newline

<Context>\{context\}</Context>\newline
<Text>\{text\}</Text>\newline
Return ONLY a list of facts, with no additional text." & Decompose response \\
    \hline
    "Given context and a claim, generate one specific, clear question that has its answer contained in the claim. The generated question must be self-contained and related to the context.
Return only the question, with no additional text.\newline

Context: \{context\}\newline
Claim: \{claim\}" & Claim-level questions \\
    \hline
    "Answer the following question based on the given context. Format your answer in one sentence:\newline

Context: \{context\}\newline
Question: \{question\}\newline

Answer: " & Question answering \\
    \hline
    "You will be given a statement and a context. Please estimate how much of the context contradicts the statement?
Your final answer should be a percentage number between 0 and 100, representing the percentage of the context that contradicts the statement.\newline

<Statement>\newline
\{statement\}\newline
</Statement>\newline

<Context>\newline
\{context\}\newline
</Context>\newline

Return your answer as a percentage number ONLY, with no additional text." & Claim-level faithfulness \\
    \hline
    "Is the following claim correct according to the reference passage? Choose your answer from <correct/incorrect/not\_enough\_information>. \newline

<Claim>{claim}</Claim> \newline

<Reference>{reference}</Reference>" & Evaluate correctness \\
    \hline
  \end{tabular}
  }
  \caption{Prompts used in IUQ.}
  \label{tab:pipeline_prompts}
\end{table*}

\end{document}